\documentclass[11pt]{article}
\usepackage{acl2016}
\usepackage{times}
\usepackage{url}
\usepackage{latexsym}
\usepackage{amsmath}
\usepackage{graphicx}  
\usepackage{bigfoot}
\usepackage{algorithm2e}
\usepackage{multirow}

\aclfinalcopy 

\title{Domain Adaptation for Neural Networks by Parameter Augmentation}

\author{Yusuke Watanabe \\
  SONY \\ 
  1-7-1 Konan Minato-ku,\\  Tokyo, Japan  \\
  {\tt \small YusukeB.Watanabe@jp.sony.com} \\\And
  Kazuma Hashimoto,   Yoshimasa Tsuruoka \\
  University of Tokyo \\
  7-3-1 Hongo, Bunkyo-ku, \\ Tokyo, Japan \\
  {\tt \small hassy@logos.t.u-tokyo.ac.jp} \\
  {\tt \small tsuruoka@logos.t.u-tokyo.ac.jp} \\
}

\date{}

\begin{document}
\maketitle
\begin{abstract}
We propose a simple domain adaptation method for neural networks in a supervised setting.
Supervised domain adaptation is a way of improving the generalization performance on
the target domain by using the source domain dataset, assuming that both of the datasets are labeled.
Recently, recurrent neural networks have been shown to be successful on a variety of NLP tasks
such as caption generation; however, the existing domain adaptation techniques are
limited to (1) tune the model parameters by the target dataset after the training by the source dataset,
or (2) design the network to have dual output, one for the source domain and the other for the target domain.
Reformulating the idea of the domain adaptation technique proposed by \newcite{daume:07},
we propose a simple domain adaptation method,
which can be applied to neural networks trained with a cross-entropy loss.
On captioning datasets, we show performance improvements over other domain adaptation methods.
\end{abstract}

\section{Introduction}
Domain adaptation is a machine learning paradigm
that aims at improving the generalization performance 
of a new (target) domain
by using a dataset from the original (source) domain.
Suppose that, as the source domain dataset,
we have a captioning corpus,
consisting of images of daily lives and
each image has captions.
Suppose also that we would like to generate captions for exotic cuisine, 
which are rare in the corpus.
It is usually very costly to make a new corpus for the target domain, i.e.,
taking and captioning those images.
The research question here is how we can leverage the source domain dataset
to improve the performance on the target domain.

As described by Daum{\'e}~\shortcite{daume:07}, 
there are mainly two settings of domain adaptation:
fully supervised and semi-supervised.
Our focus is the supervised setting, where
both of the source and target domain datasets are labeled.
We would like to use the label information of the source domain
to improve the performance on the target domain.

Recently, Recurrent Neural Networks (RNNs)
have been successfully applied to various tasks
in the field of natural language processing (NLP),
including language modeling~\cite{mikolov2010recurrent}, 
caption generation~\cite{vinyals2015show} and
parsing~\cite{vinyals2015grammar}.

For neural networks, there are two standard methods for supervised domain adaptation~\cite{mou2016transferable}.
The first method is {\it fine tuning}:
we first train the model with the source dataset and
then tune it with the target domain dataset~\cite{venugopalan2014translating,kim2014convolutional}.
Since the objective function of neural network training is non-convex,
the performance of the trained model can depend on the initialization of the parameters.
This is in contrast with the convex methods such as Support Vector Machines (SVMs).
We expect that the first training gives a good initialization of the parameters, 
and therefore the latter training gives a good generalization even if the 
target domain dataset is small.
The downside of this approach is the lack of the optimization objective.

The other method is to design the neural network so that it has two outputs.
The first output is trained with the source dataset and the other output is
trained with the target dataset, where the input part is shared among the domains.
We call this method {\it dual outputs}.
This type of network architecture has been successfully applied to 
multi-task learning in NLP such as part-of-speech tagging and 
named-entity recognition~\cite{collobert2011scratch,yang2016multitask}.
                                                 
In the NLP community, there has been a large body of previous work on domain adaptation. 
One of the state-of-the-art methods for the supervised domain adaptation
is {\it feature augmentation}~\cite{daume:07}.
The central idea of this method is to augment the original features/parameters in order to model the
source specific, target specific and general behaviors of the data.
However, it is not straight-forward to apply it to neural network models 
in which the cost function has a form of log probabilities.

In this paper, we propose a new domain adaptation method for neural networks.
We reformulate the method of \newcite{daume:07} 
and derive an objective function using convexity of the loss function.
From a high-level perspective,
this method shares the idea of feature augmentation.
We use redundant parameters for the source, target and general domains, where the general 
parameters are tuned to model the common characteristics of the datasets
and the source/target parameters are tuned for domain specific aspects.

In the latter part of this paper, 
we apply our domain adaptation method to a neural captioning model
and show performance improvement over other standard methods on several datasets and metrics.
In the datasets, the source and target have different word distributions,
and thus adaptation of output parameters is important.
We augment the output parameters to facilitate adaptation.
Although we use captioning models in the experiments,
our method can be applied to any neural networks trained with a cross-entropy loss.

\section{Related Work}
There are several recent studies applying domain adaptation 
methods to deep neural networks. However, few studies have focused on
improving the fine tuning and dual outputs methods in the supervised setting.

\newcite{sun2015return} have proposed an unsupervised domain adaptation method
and apply it to the features from deep neural networks.
Their idea is to minimize the domain shift by aligning the second-order statistics of 
source and target distributions.
In our setting, it is not necessarily true that
there is a correspondence between the source and target input distributions,
and therefore we cannot expect their method to work well.

\newcite{wen2016multi} have proposed a procedure to
generate natural language for multiple domains of spoken dialogue systems.
They improve the fine tuning method by pre-trainig with synthesized data. However,
the synthesis protocol is only applicable to the spoken dialogue system.
In this paper, we focus on domain adaptation methods 
which can be applied without dataset-specific tricks.

\newcite{yang2016multitask} have conducted a series of experiments
to investigate the transferability of neural networks for NLP.
They compare the performance of two transfer methods called {\tt INIT} and {\tt MULT},
which correspond to the fine tuning and dual outputs methods in our terms.
They conclude that {\tt MULT} is slightly better than or comparable to {\tt INIT};
this is consistent with our experiments shown in section~\ref{sec:experiments}.
Although they obtain little improvement by transferring the output parameters,
we achieve significant improvement by augmenting parameters in the output layers.

\section{Domain adaptation and language generation}
We start with the basic notations and formalization for domain adaptation.
Let $\mathcal{X}$ be the set of inputs and $\mathcal{Y}$ be the outputs.
We have a source domain dataset $D^s$, which is sampled from some distribution $\mathcal{D}^s$.
Also, we have a target domain dataset $D^t$, which is sampled from 
another distribution $\mathcal{D}^t$.
Since we are considering supervised settings,
each element of the datasets has a form of input output pair $(x,y)$.
The goal of domain adaptation is to learn a function 
$f : \mathcal{X} \rightarrow \mathcal{Y}$ that models the
input-output relation of $D^t$.
We implicitly assume that there is a connection between the source and target distributions
and thus can leverage the information of the source domain dataset.
In the case of image caption generation,
the input $x$ is an image (or the feature vector of an image) and 
$y$ is the caption (a sequence of words).

In language generation tasks, a sequence of words is generated
from an input $x$.
A state-of-the-art model for language generation is 
LSTM (Long Short Term Memory) initialized by a context vector
computed by the input \cite{vinyals2015show}.
LSTM is a particular form of recurrent neural network, 
which has three gates and a memory cell.
For each time step $t$, 
the vectors $c_t$ and $h_t$ are computed from 
$u_t, c_{t-1}$ and $h_{t-1}$ by the following equations:
\begin{align*}
&i = \sigma(W_{ix} u_t + W_{ih} h_{t-1})  \\
&f = \sigma(W_{fx} u_t + W_{fh} h_{t-1})  \\
&o = \sigma(W_{ox} u_t + W_{oh} h_{t-1})  \\
&g = \tanh(W_{gx} u_t + W_{gh} h_{t-1})  \\
&c_t = f \odot c_{t-1} + i \odot g    \\
&h_t = o \odot \tanh(c_t),
\end{align*}
where $\sigma$ is the sigmoid function and $\odot$ is the element-wise product.
Note that all the vectors in the equations have the same dimension $n$, called the cell size.
The probability of the output word at the $t$-th step, $y_t$, is computed by 
\begin{equation}
 p(y_t|y_1,\ldots,y_{t-1},x) = {\rm Softmax}(W h_t),  \label{eq:softmax}
\end{equation}
where $W$ is a matrix with a size of vocabulary size times $n$.
We call this matrix as the parameter of the {\it output layer}.
The input $u_t$ is given by the word embedding of $y_{t-1}$.

\begin{figure}[t]
  \centering
  \includegraphics[bb=0 0 456.6 291.6,width=7.5cm]{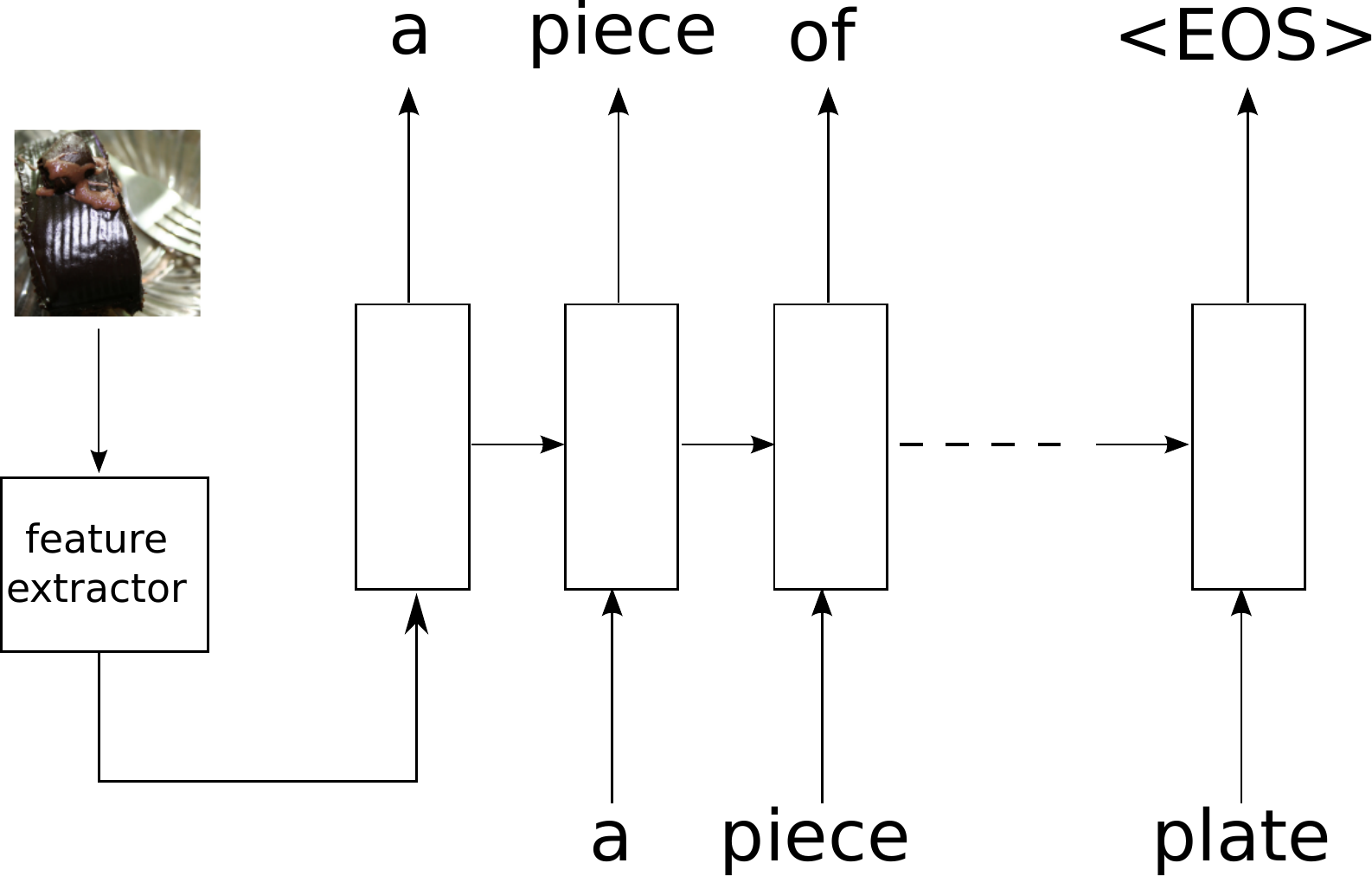}    
  \caption{A schematic view of the LSTM captioning model. 
   The first input to the LSTM is an image feature.
   Then a sentence ``{\it a piece of chocolate cake that is on a glass plate}'' is generated.
   The generation process ends with the EOS symbol.
   } \label{fig_caption_generation}
\end{figure}

To generate a caption, 
we first compute feature vectors of the image, and put it into the beginning of the LSTM as 
\begin{equation}
 u_{0} = W_{0} {\rm CNN}(x),
\end{equation}
where $W_0$ is a tunable parameter matrix and
${\rm CNN}$ is a feature extractor usually given by a convolutional neural network.
Output words, $y_t$, are selected in order and
each caption ends with special symbol \verb|<EOS>|.
The process is illustrated in Figure~\ref{fig_caption_generation}.
Note that the cost function for the generated caption is
\begin{equation*}
   \log p(y|x) = \sum_{t} \log p(y_t|y_1,\ldots,y_{t-1}, x),
\end{equation*}
where the conditional distributions are given by Eq.~(\ref{eq:softmax}).
The parameters of the model are optimized to minimize the cost on the training dataset.
We also note that there are extensions of the models with attentions
\cite{xu2015show,bahdanau2014neural}, but 
the forms of the cost functions are the same.

\section{Domain adaptation for language generation}
In this section, we review standard domain adaptation techniques
which are applicable to the neural language generation.
The performance of these methods is compared in the next section.

\subsection{Standard and baseline methods}
A trivial method of domain adaptation is simply ignoring the source dataset,
and train the model using only the target dataset.
This method is hereafter denoted by {\sc TgtOnly}.
This is a baseline and any meaningful method must beat it.

Another trivial method is {\sc SrcOnly}, where only 
the source dataset is used for the training.
Typically, the source dataset is bigger than that of the target,
and this method sometimes works better than {\sc TgtOnly}.

Another method is {\sc All}, 
in which the source and target datasets are combined and used for the training.
Although this method uses all the data,
the training criteria enforce the model to perform well on both of the domains,
and therefore the performance on the target domain is not necessarily high.
\footnote{
As a variant of this method, we can weight the samples in each domain.
This is a kind of interpolation between {\sc TgtOnly}, {\sc All} and {\sc SrcOnly}.
We do not consider this method in the latter experiments because 
we observe little improvement over {\sc All}.
}

An approach widely used in the neural network community is {\sc FineTune}.
We first train the model with the source dataset and then it is
used as the initial parameters for training the model with the target dataset.
The training process is stopped in reference to the development set
in order to avoid over-fitting.
We could extend this method by posing a regularization term (e.g. $l_2$ regularization)
in order not to deviate from the pre-trained parameter. 
In the latter experiments, however, we do not pursue this direction because 
we found no performance gain. 
Note that it is hard to control the scales of the regularization
for each part of the neural net
because there are many parameters having different roles.

Another common approach for neural domain adaptation is {\sc Dual}.
In this method, the output of the network is ``dualized''. 
In other words, we use different parameters $W$ in Eq.~(\ref{eq:softmax}) for the source and target domains.
For the source dataset, the model is trained with the first output and the second for the target dataset.
The rest of the parameters are shared among the domains.
This type of network design is often used for multi-task learning.

\subsection{Revisiting the feature augmentation method}
Before proceeding to our new method,
we describe the feature augmentation method \cite{daume:07} from our perspective.
let us start with the feature augmentation method.

Here we consider the domain adaptation of a binary classification problem.
Suppose that we train SVM models for the source and target domains separately. 
The objective functions have the form of
\begin{align*}
\frac{1}{n_s} \sum_{(x,y) \in \mathcal{D}_s}
 \max (0, 1 - y(w_s^T \Phi(x))) + \lambda \Vert w_s \Vert^2  \\
\frac{1}{n_t} \sum_{(x,y) \in \mathcal{D}_t}
 \max (0, 1 - y(w_t^T \Phi(x))) + \lambda \Vert w_t \Vert^2 ,
\end{align*}
where $\Phi(x)$ is the feature vector and $w_s, w_t$ are the SVM parameters.
In the feature augmentation method, 
the parameters are decomposed to  $w_s = \theta_g + \theta_s$ and $w_t = \theta_g + \theta_t$.
The optimization objective is different from the sum of the above functions:
\begin{align*}
& \frac{1}{n_s} 
 \sum_{(x,y) \in \mathcal{D}_s}  \max (0, 1 - y(w_s^T \Phi(x))) \\
&+\lambda (\Vert \theta_g \Vert^2 + \Vert \theta_s \Vert^2 )   \\
&+ \frac{1}{n_t} \sum_{(x,y) \in \mathcal{D}_t}  \max (0, 1 - y(w_t^T \Phi(x))) \\
&+ \lambda (\Vert \theta_g \Vert^2 + \Vert \theta_t \Vert^2 ),
\end{align*}
where the quadratic regularization terms
$\Vert \theta_g + \theta_s \Vert^2$ and
$\Vert \theta_g + \theta_t \Vert^2$
are changed to 
$\Vert \theta_g \Vert^2 + \Vert \theta_s \Vert^2$ 
and
$\Vert \theta_g \Vert^2 + \Vert \theta_t \Vert^2$, respectively.
Since the parameters $\theta_g$ are shared, we cannot optimize the problems separately.

This change of the objective function can be understood as 
adding additional regularization terms
\begin{align*}
 2(\Vert \theta_g \Vert^2 + \Vert \theta_t \Vert^2  ) - \Vert \theta_g + \theta_t \Vert^2,  \\
 2(\Vert \theta_g \Vert^2 + \Vert \theta_s \Vert^2  ) - \Vert \theta_g + \theta_s \Vert^2.
\end{align*}
We can easily see that those are equal to $\Vert \theta_g - \theta_t \Vert^2$ and
$\Vert \theta_g - \theta_s \Vert^2$, respectively and thus
this additional regularization enforces $\theta_g$ and $\theta_t$ 
(and also $\theta_g$ and $\theta_s$) not to be far away.
This is how the feature augmentation method shares the domain information 
between the parameters $w_s$ and $w_t$.

\subsection{Proposed method}
Although the above formalization is for an SVM, which has the quadratic cost of parameters, 
we can apply the idea to the log probability case.

In the case of RNN language generation,
the loss function of each output is a cross entropy applied to the softmax output
\begin{align}
  -\log & p_s(y|y_1, \ldots, y_{t-1}, x)  \nonumber \\
    &= -w_{s,y}^T h + \log Z(w_s;h),  \label{eq:source_loss}
\end{align}
where $Z$ is the partition function and
$h$ is the hidden state of the LSTM computed by $y_0, \ldots, y_{t-1}$ and $x$.
Again we decompose the word output parameter as $w_s = \theta_g + \theta_s$. 
Since $\log Z$ is convex with respect to $w_s$,
we can easily show that the Eq.~(\ref{eq:source_loss}) is bounded above by
\begin{align*}
-&\theta_{g,y}^T h + \frac{1}{2} \log Z(2 \theta_g;x)   \\
&-\theta_{s,y}^T h +\frac{1}{2}  \log Z(2 \theta_s;x).
\end{align*}
The equality holds if and only if $\theta_g = \theta_s$.
Therefore, optimizing this upper-bound effectively enforces the parameters to be close
as well as reducing the cost.

The exact same story can be applied to the target parameter
$w_t = \theta_g + \theta_t$.
We combine the source and target cost functions and
optimize the sum of the above upper-bounds.
Then the derived objective function is
\begin{align*}
  \frac{1}{n_s} \sum_{(x,y) \in \mathcal{D}_s}   
[
  -\theta_{g,y}^T h& + \frac{1}{2} \log Z(2 \theta_g;x)  \\
  &-\theta_{s,y}^T h + \frac{1}{2} \log Z(2 \theta_s;x) 
]
\\
+ \frac{1}{n_t} \sum_{(x,y) \in \mathcal{D}_t}  
[
-\theta_{g,y}^T h &+ \frac{1}{2} \log Z(2 \theta_g;x)  \\
& -\theta_{t,y}^T h + \frac{1}{2} \log Z(2 \theta_t;x)
].
\end{align*}

If we work with the sum of the source and target versions of Eq.~(\ref{eq:source_loss}),
the method is actually the same as {\sc Dual} because the parameters $\theta_g$ is completely redundant.
The difference between this objective and the proposed upper bound works as a regularization term,
which results in a good generalization performance.

Although our formulation has the unique objective, there are three types of
cross entropy loss terms given by $\theta_g$, $\theta_s$ and $\theta_t$.
We denote them by $\ell(\theta_g), \ell(\theta_s)$ and $\ell(\theta_t)$, respectively.
For the source data, the sum of general and source loss terms is optimized,
and for the target dataset the sum of general and target loss terms is optimized.

The proposed algorithm is summarized in Algorithm~\ref{alg_proposed}.
Note that $\theta_h$ is the parameters of the LSTM except for the output part.
In one epoch of the training, we use all data once.
We can combine any parameter update methods for neural network training
such as Adam~\cite{kingma2014adam}.

\RestyleAlgo{boxruled}
\LinesNumbered
\begin{algorithm}[t]
  \caption{Proposed Method \label{alg_proposed}}
\While{{\rm True}}{
   Select a minibatch of data from source or target dataset \\
  \If{source}{
      Optimize  $\ell(\theta_g) + \ell(\theta_s)$ 
      with respect to $\theta_g, \theta_s, \theta_h$ for the minibatch
    }
  \Else{
      Optimize $\ell(\theta_g) + \ell(\theta_t)$  
      with respect to $\theta_g, \theta_t, \theta_h$ for the minibatch
    }
  \If{development error increases}{
    break\;
    }
}
 Compute $w_t = \theta_g + \theta_t$
 and $w_s = \theta_g + \theta_s$.
 Use these parameters as the output parameters for each domain.
\end{algorithm}

\section{Experiments}
\label{sec:experiments}
We have conducted domain adaptation experiments on the following three datasets.
The first experiment focuses on the situation where the domain adaptation is useful.
The second experiment show the benefit of domain adaptation for both directions:
from source to target and target to source.
The third experiment shows an improvement in another metric.
Although our method is applicable to any neural network with a cross entropy loss,
all the experiments use caption generation models 
because it is one of the most successful neural network applications in NLP.

\subsection{Adaptation to food domain captioning}
This experiment highlights a typical scenario in which domain adaptation is useful.
Suppose that we have a large dataset of captioned images, which are taken from daily lives, but
we would like to generate high quality captions for more specialized
domain images such as minor sports and exotic food.
However, captioned images for those domains are quite limited due to the annotation cost.
We use domain adaptation methods to improve the captions of the target domain.

To simulate the scenario, we split the Microsoft COCO dataset into food and non-food domain datasets.
The MS COCO dataset contains approximately 80K images for training and
40K images for validation; each image has 5 captions~\cite{lin2014microsoft}.
The dataset contains images of diverse categories, including
animals, indoor scenes, sports, and foods.
We selected the ``food category'' data by scoring the captions
according to how much those are related to the food category.
The score is computed based on wordnet similarities~\cite{miller1995wordnet}.
The training and validation datasets are split by the score with the same threshold.
Consequently, the food dataset has 3,806 images for training and 1,775 for validation.
The non-food dataset has 78,976 images for training and 38,749 for validation.

The selected pictures from the food domain are 
typically a close-up of foods or people eating some foods.
Table~\ref{table_foodcaption} shows some captions from the food
and non-food domain datasets.
Table~\ref{table_topwords} shows the top twenty frequent words in the two datasets
except for the stop words.
We observe that the frequent words are largely different,
but still there are some words common in both datasets.

\begin{table}[t]
\bgroup
\def\arraystretch{1.4}
\small
\centering
\begin{tabular}{|l|}
\hline
Closeup of bins of food that include broccoli and bread.  \\
A woman sitting in front of a table with a plate of food.  \\
A large pizza covered in cheese and toppings. \\
\hline
People shopping in an open market for vegetables. \\
A purse sits at the foot of one of the large beds.  \\
A large television screen in a large room. \\
\hline
\end{tabular}
  \caption{Examples of annotated captions from food domain dataset (top) and
  non-food dataset (bottom).} \label{table_foodcaption}
\egroup
\end{table}

\begin{table}[t]
\small
\centering
\begin{tabular}{|l|l|}
\hline
food &
food, plate, table, pizza, sitting, man, white, \\
&  two, eating, people, sandwich, woman, next, \\
&  plates, vegetables, cheese, bowl,   \\
\hline
non-food &  man, sitting, two, standing, people, next, white, \\
&  woman, street, holding, person, table, large, \\
&  down, top, group, field, tennis, small, near, \\
\hline
\end{tabular}
  \caption{Top twenty frequent words from the food/non-food datasets.} \label{table_topwords}
\end{table}

To model the image captaining, we use LSTMs as described in the previous section.
The image features are computed by the trained GoogLeNet and 
all the LSTMs have a single layer with 300 hidden units~\cite{szegedy2015going}.
We use a standard optimization method, Adam~\cite{kingma2014adam}
with hyper parameters $\alpha=0.001$, $\beta_1=0.9$ and $\beta_2=0.999$.
We stop the training based on the loss on the development set.
After the training we generate captions by beam search, where
the size of the beam is 5.
These settings are the same in the latter experiments.

We compare the proposed method with other baseline methods.
For all the methods, we use Adam with the same hyper parameters.
In {\sc FineTune}, we did not freeze any parameters during the target training.
In {\sc Dual}, all samples in source and target datasets are weighted equally.

We evaluated the performance of the domain adaptation methods 
by the qualities of the generated captions.
We used BLEU, METOR and CIDEr scores for the evaluation.
The results are summarized in Table~\ref{table_food}.
\footnote{We use scripts in \verb|https://github.com/tylin/| \verb|coco-caption|
to compute BLEU, METEOR and CIDEr scores.}
We see that the proposed method improves in most of the metrics.
The baseline methods {\sc SrcOnly} and {\sc TgtOnly} are worse than other methods,
because they use limited data for the training.
Note that the CIDEr scores correlate with human evaluations better than BLEU and METOR scores~\cite{vedantam2015cider}.

Generated captions for sample images are shown in Table~\ref{table_decoded}.
In the first example, {\sc All} fails to identify the chocolate cake because
there are birds in the source dataset which somehow look similar to chocolate cake.
We argue that {\sc Proposed} learns birds by the source parameters and 
chocolate cakes by the target parameters, and thus succeeded in generating appropriate captions.

\begin{table}[t]
\small
\centering
\begin{tabular}{|c|cccc|c|c|}
\hline
  &    B1 & B2 & B3 &  B4 & M &  C \\
\hline
 {\sc SrcOnly}  & 60.4 &  42.3  & 30.6 & 21.2 &  19.4 &  36.4 \\
 {\sc TgtOnly}  & 63.0 &  45.5  & 33.0 & 24.0 &  20.9 &  35.8 \\
 {\sc All}      & 61.0 &  45.1  & 32.7 & 23.7 & 20.2 &  39.9 \\
 {\sc FineTune} & 61.9 &  45.8  & 33.6 & 24.6 &  21.5 &  39.8 \\
 {\sc Dual}     & {\bf 63.3} &  46.3  & 33.7 & {\bf 24.7} &  21.2 &  40.7 \\
 {\sc Proposed} & 63.2 &  {\bf 46.8}  & {\bf 34.0} & {\bf 24.7} &  {\bf 21.7} &  {\bf 42.8} \\
\hline
\end{tabular}
  \caption{Results of the domain adaptation to the food dataset.
The evaluation metrics are BLEU, METOR and CIDEr.
The proposed method is the best in most of the metrics.} \label{table_food}
\end{table}

%



\begin{table*}[ht] 
  \begin{tabular}
      {|l|l|l|}  
      \hline
      \multirow{7}{*}
      {\includegraphics[bb=0 0 150 150,width=2.4cm]{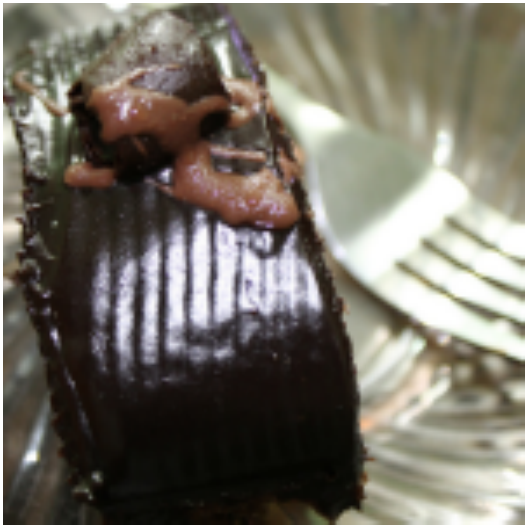}} 
      & True Caption &    A piece of chocolate cake that is on a glass plate.   \\ \cline{2-3}
      & {\sc SrcOnly} &   a bird sitting on top of a tree branch    \\
      & {\sc TgtOnly} &   a piece of chocolate cake sitting on a white plate    \\  
      & {\sc All} &   a close up of a bird on a tree branch    \\  
      & {\sc FineTune} &   a close up of a plate of food on a plate     \\  
      & {\sc Dual} &   a close up of a plate of food     \\        
      & {\sc Proposed} &   {\bf a close up of a piece of chocolate cake on a plate }   \\  
      \hline
      \hline 
      \multirow{7}{*}
      {\includegraphics[bb=0 0 150 150,width=2.4cm]{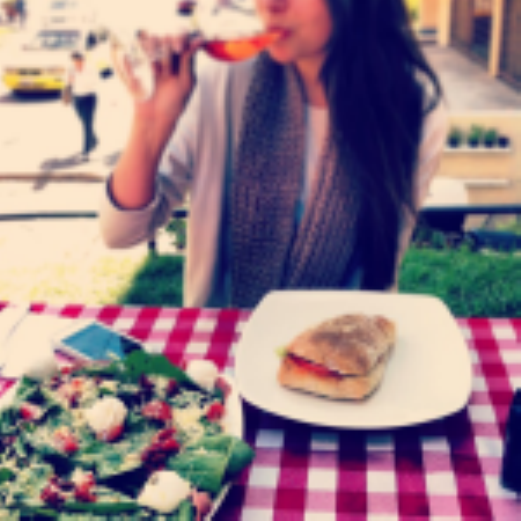}} 
      & True Caption &  The woman with a sandwich on her plate is drinking from a wine glass.     \\ \cline{2-3}
      & {\sc SrcOnly} &   a woman holding a cake on a table    \\
      & {\sc TgtOnly} &    a woman is eating a slice of pizza   \\  
      & {\sc All} &  a person holding a cake on a plate with a fork    \\  
      & {\sc FineTune} &   a close up of a plate of food on a table   \\  
      & {\sc Dual} &   a group of people sitting at a table eating food     \\              
      & {\sc Proposed} &   {\bf a woman sitting at a table with a plate of food}   \\  
      \hline
  \end{tabular}
  \caption{Examples of generated captions for food dataset images.}  \label{table_decoded}
\end{table*}


\subsection{Adaptation between MS COCO and Flickr30K}
In this experiment, we explore the benefit of adaptation from both sides of the domains.
Flickr30K is another captioning dataset, consisting of 30K images, and each image has five captions~\cite{young2014image}.
Although the formats of the datasets are almost the same,
the model trained by the MS COCO dataset does not work well for the Flickr 30K dataset and vice versa.
The word distributions of the captions are considerably different.
If we ignore words with less than 30 counts, MS COCO has 3,655 words and Flicker30K has 2732 words;
and only 1,486 words are shared. 
Also, the average lengths of captions are different.
The average length of captions in Flickr30K is 12.3
while that of MS COCO is 10.5.

The first result is the domain adaptation from MS COCO to Flickr30K,
summarized in Table~\ref{table_to_flickr}.
Again, we observe that the proposed method achieves the best score among the other methods.
The difference between {\sc All} and {\sc FineTune} is bigger than in the previous setting
because two datasets have different captions even for similar images.
The scores of {\sc FineTune} and {\sc Dual} are at almost the same level.

\begin{table}[t]
\small
\centering
\begin{tabular}{|c|cccc|c|c|}
\hline
  &    B1 & B2 & B3 &  B4 & M &  C \\
\hline
 {\sc SrcOnly}  &  50.8 & 30.5 & 18.6 & 11.5 &  12.9 &  16.0 \\
 {\sc TgtOnly}  &  52.8 & 34.6 & 22.8 & 15.3 &  15.7 &  22.5 \\
 {\sc All}      &  52.6 & 34.4 & 22.5 & 14.8 &  15.6 &  23.1 \\
 {\sc FineTune} &  55.2 & 36.6 & 24.3 & 16.3 &  16.0 &  26.2 \\
 {\sc Dual}     &  56.0 & 36.9 & 24.1 & 16.0 &  15.9 &  25.8 \\
 {\sc Proposed} &  {\bf 56.7} & {\bf 37.8} & {\bf 25.5} & {\bf 17.4} &  {\bf 16.1} &  {\bf 27.9} \\
\hline
\end{tabular}
 \caption{\label{table_to_flickr} Domain adaptation from MSCOCO to Flickr30K dataset.}
\end{table}

The second result is the domain adaptation from Flickr30K to MS COCO shown in Table~\ref{table_to_mscoco}.
This may not be a typical situation because the number of samples in the target domain
is larger than that of the source domain.
The {\sc SrcOnly} model is trained only with Flickr30K and tested on the MS COCO dataset.
We observe that {\sc FineTune} gives little benefit over {\sc TgtOnly}, which implies that
the difference of the initial parameters has little effect in this case.
Also, {\sc Dual} gives little benefit over {\sc TgtOnly},
meaning that the parameter sharing except for the output layer is not important in this case.
Note that the CIDEr score of {\sc Proposed} is slightly improved.

\begin{table}[t]
\small
\centering
\begin{tabular}{|c|cccc|c|c|}
\hline
  &    B1 & B2 & B3 &  B4 & M &  C \\
\hline
 {\sc SrcOnly}  &  44.0 &  25.1 &  14.1 & 8.6 & 13.5  &  15.5 \\
 {\sc TgtOnly}  &  64.0 &  45.9 &  32.6 & 23.2 & 21.0  & 70.2  \\
 {\sc All}      &  63.0 &  44.9 & 31.4 & 22.2 & 21.0  & 67.4  \\
 {\sc FineTune} &  63.6 & 45.7 & 32.7 & {\bf 23.5} & 20.9  & 70.5  \\
 {\sc Dual} &  {\bf 65.0} & {\bf 46.6} & 32.8 & 23.1 & 21.0  & 70.3  \\ 
 {\sc Proposed} &  64.3 & 46.5 & {\bf 33.0} & 23.4 &  {\bf 21.1} &  {\bf 71.0} \\
\hline
\end{tabular}
 \caption{\label{table_to_mscoco} Domain adaptation from Flickr30K to MSCOCO dataset.}
\end{table}

\begin{figure}[t]
 \includegraphics[bb=0 0 600 240,width=10cm]{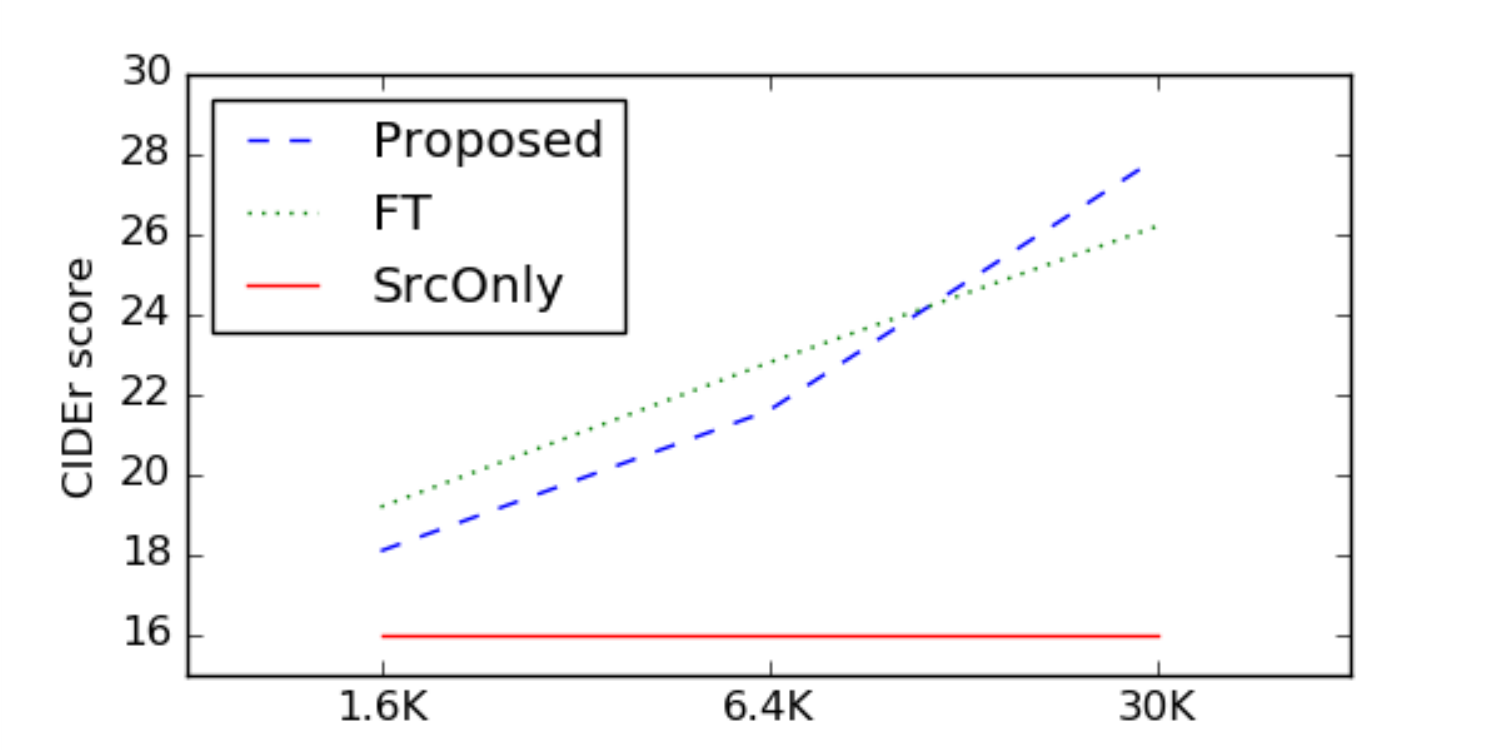}
  \caption{Comparison of CIDEr score of {\sc FineTune} and {\sc Proposed}} \label{fig001}
\end{figure}

Figure~\ref{fig001} shows the comparison of {\sc FineTune} and {\sc Proposed},
changing the number of the Flickr samples to 1600, 6400 and 30K.
We observe that {\sc FineTune} works relatively well when the target domain dataset is small.

\subsection{Answer sentence selection}
In this experiment, we use the captioning model as an affinity measure of images and sentences. 
TOEIC part 1 test consists of four-choice questions for English learners.
The correct choice is the sentence that best describes the shown image.
Questions are not easy because there are confusing keywords in wrong choices.
An example of the question is shown in Table~\ref{table_toeic1}.
We downloaded 610 questions from \verb|http://www.english-test.net/toeic/| \verb|listening/|.

\begin{table*}[ht] 
\bgroup
\def\arraystretch{1.4}
\centering
  \begin{tabular}
      {|l|l|}  
      \hline
      \multirow{5}{*}
      {\includegraphics[bb=0 0 305 225,width=4cm]{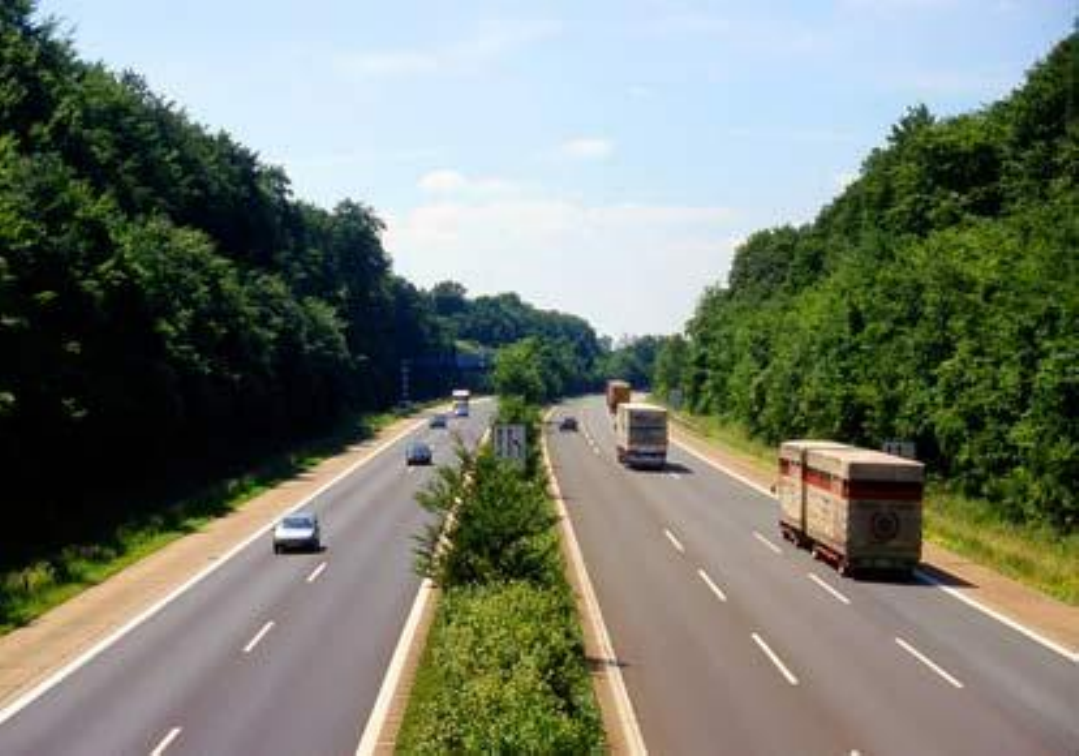}} 
      & (A) Traffic is building up on the motorway.    \\
      & (B) There are more lorries on this motorway than cars.    \\  
      & (C) Traffic is flowing freely on the motorway.    \\    
      & (D) The vehicles are traveling too close to one another on the motorway.     \\  
      & \\
      \hline
  \end{tabular}
  \caption{A sample question from TOEIC part 1 test. The correct answer is (C).}  \label{table_toeic1}
\egroup
\end{table*}

Our approach here is to select the most probable choice given the image by captioning models.
We train captioning models with the images and correct answers from the training set.
Since the TOEIC dataset is small, domain adaptation can give a large benefit.
We compared the domain adaptation methods by the percentage of correct answers.
The source dataset is 40K samples from MS COCO and the target dataset is the TOEIC dataset. 
We split the TOEIC dataset to 400 samples for training and 210 samples for testing.

The percentages of correct answers for each method are summarized in Table~\ref{table_toeicscore}.
Since the questions have four choices, all methods should perform better than 25\%.
{\sc TgtOnly} is close to the baseline because the model is trained with only 400 samples.
As in the previous experiments, {\sc FineTune} and {\sc Dual} are better than {\sc All} and
{\sc Proposed} is better than the other methods.

\begin{table}[t]
\centering
\begin{tabular}{|c|r|}
\hline
  &  correct answer\\
\hline
 {\sc SrcOnly}  &  29.1\% \\
 {\sc TgtOnly}  &  28.1\% \\
 {\sc All}      &  31.0\% \\
 {\sc FineTune} &  33.3\% \\
 {\sc Dual} &  33.3\% \\
 {\sc Proposed} &  {\bf 35.7\%} \\
\hline
\end{tabular}
 \caption{Domain adaptation to TOEIC dataset.}
 \label{table_toeicscore} 
\end{table}

\newpage

\section{Conclusion and Future Work}
We have proposed a new method for supervised domain adaptation of neural networks.
On captioning datasets, we have shown that the method outperforms other standard adaptation methods
applicable to neural networks.

The proposed method only decomposes the output word parameters,
where other parameters, such as word embedding, are completely shared across the domains.
Augmentation of parameters in the other part of the network 
would be an interesting direction of future work.

\bibliography{myref}
\bibliographystyle{acl2016}

\end{document}